# FINGERTIP IN THE EYE: A CASCADED CNN PIPELINE FOR THE REAL-TIME FINGERTIP DETECTION IN EGOCENTRIC VIDEOS

*Xiaorui Liu, Yichao Huang, Xin Zhang\*, Lianwen Jin\**

School of Electronic and Information Engineering
South China University of Technology
\*Email:  eexinzhang@scut.edu.cn, eelwjin@scut.edu.cn

**ABSTRACT**

**We introduce a new pipeline for hand localization and fingertip detection. For RGB images captured from an egocentric vision mobile camera, hand and fingertip detection remains a challenging problem due to factors like background complexity and hand shape variety. To address these issues accurately and robustly, we build a large scale dataset named Ego-Fingertip and propose a bi-level cascaded pipeline of convolutional neural networks, namely, Attention-based Hand Detector as well as Multi-point Fingertip Detector. The proposed method significantly tackles challenges and achieves satisfactorily accurate prediction and real-time performance compared to previous hand and fingertip detection methods.**

## 1. INTRODUCTION

Nature human-computer interaction (NHCI) via smart head-mounted devices, such as Microsoft HoloLens [1] and Google Glass [2], reveals the potential for the novel HCI pattern like gesture interaction based on the hand and fingertip. Nevertheless, the accurate robust hand and fingertip detection remains challenging problem in the dynamic scene taken by an egocentric RGB mobile camera.

Current fingertip detection researches mainly use hand-crafted features and mostly test on the limited data of indoor environment with clean background. However, the real-world images of dynamic scenarios impose great challenges for algorithms using traditional vision features, such as skin-color model [3], background model [4] and RGB-D model [5] etc. CNN-based methods show potential, and a few latest works based on CNN have shown clear improvements on the object detection, such as Overfeat [6], R-CNN [7], SPP [8], Fast R-CNN [9] and Faster R-CNN [10]. To achieve real-time detection, Redmon proposed YOLO [24], which redefines detection as a regression problem. Besides, Sun's method [11] on facial landmark detection and Toshev's approach [12] on body joint detection make significant contributions on CNN-based key point detection. Furthermore, our previous work [13] firstly applied CNN framework for the egocentric hand and fingertip detection, but its hand detection involves large error that leads to inaccurate fingertip detection.

In this paper, we apply a bi-level cascaded CNN pipeline, as Figure 1 shows, which is beneficial to filter complex background information. An attention-based CNN is designed for hand detection cultivated by visual attention mechanism, and a multi-point fingertip detector based on spatial constraint is presented. Last but not the least, we establish a large-scale dataset, named *Ego-Fingertip*, for egocentric vision hand and fingertip detection. The cascaded pipeline proposed performs satisfyingly on the dataset in real time, achieving a high average overlap rate of 0.71 for hand detection and average Euclidean error of 15.71 pixels for fingertip out of 640*480 video sequences.

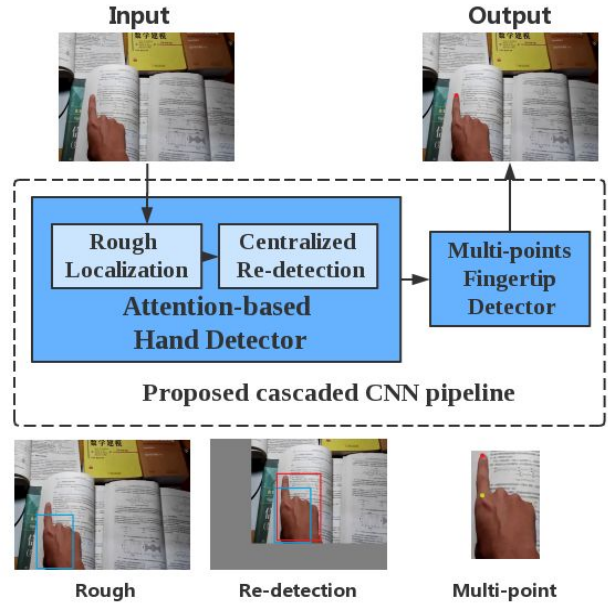

Figure 1. The cascaded finger detection pipeline with corresponding sample results

## 2. DATASET: EGO-FINGERTIP

According to Russakovsky [26], the CNN needs large training dataset to achieve satisfied feature representation learning. Hence, we establish a dataset collecting the large amount of fingertip poses, gestures and interaction process in the egocentric videos, named *Ego-Fingertip\*,* which will be released for academic research.

## 2.1. Ego-fingertip dataset detail

The *Ego-fingertip* dataset contains 93,729 RGB frames of egocentric videos captured, collected and labeled by 24 volunteers in various scenarios. Besides, the location labels for index fingertip, index finger joint and little finger are included. Correspondingly, it involves challenges from several aspects as follows: background complexity, including scene variety, illumination change, skin-like object occurrence; hand foreground diversity, including varying skin color, deformable hand shape and even motion blur. Figure 2 shows samples of the dataset.

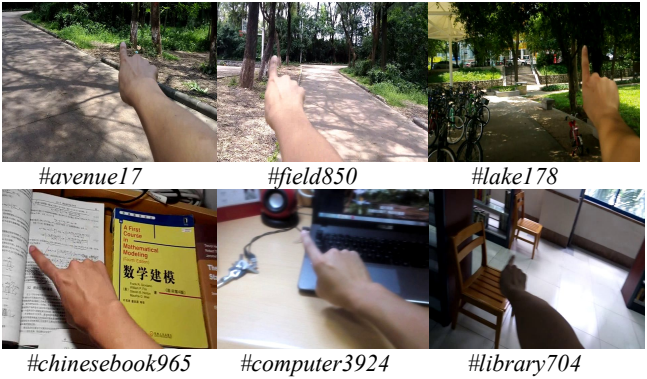

*#avenue17*　　　*#field850*　　　*#lake178*

*#chinesebook965*　　*#computer3924*　　*#library704*

Figure 2. Sample frames of the dataset.

Statistic shows that fingertip are pointing at various angle, that pointing at left occupies about 60% and pointing to up and right side occupies about 40% . When roughly define a threshold of light channel, about 30% frames are relatively dark comparing with about 70% bright samples. Moreover, skin-color-like object occurs in each packages.

## 2.2. Hand Location Analysis

We study the hand location distribution by frequency calculation of the dataset and the result is shown in Figure 3, which fits the Gaussian distribution. It is clear that the hand locates in the center more often. This conclusion matches human eye visual attention mechanism of "center bias" [14, 15], which brings inspiration for the attention-based re-detection in the following AHD.

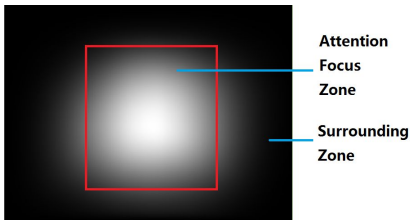

Figure 3. The hand location distribution in the dataset. The lighter color indicates the higher frequency. The Attention Focus Zone and Surrounding Zone are defined accordingly.

## 3. AHD-MFD CASCADED CNN PIPELINE

We present a bi-level cascaded CNN pipeline, as described in Figure 1, which we found is essential and effective for fingertip detection because the hand detection procedure helps to filter most part of useless complicated background.

To begin with, we discuss first level attention-based CNN, namely, Attention-based Hand Detector (AHD), and the second level CNN, namely, Multi-point Fingertip Detector (MFD). AHD focuses on detecting hand bounding box and therefore provides sub-region for MFD, which will finally locate the coordinates of fingertip.

## 3.1. Attention-based Hand Detector: AHD

### 3.1.1 Rough hand detection

We firstly apply a 5-layer CNN as feature extractor to produce feature maps, followed by a 3-layer fully connected layer as regressor to generate the left-top and right-bottom points of hand bounding box, shown in Figure 4.

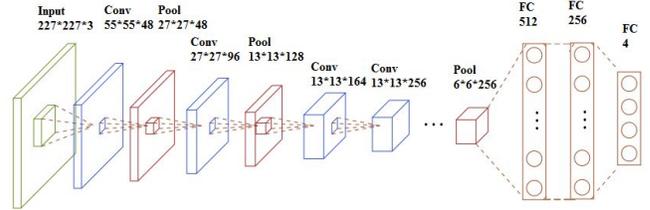

Figure 4: Hand detection network

In detail, hand features such as skin color [16], hand edge [17], their hierarchical combinations and even semantic information [18, 19], are extracted by the convolutional layers. These features represent the information of hand size and location. Following this, the fully connected layer, considered as regressor, will map those features to the left-top and right-bottom points' coordinate of the hand bounding box.

### 3.1.2 Attention-based hand re-detection

However, after analyzing the hand detection error, we found that hand detection for samples whose hand locates in Attention Focus Zone is a lot more accurate in average than in Surrounding Zone.

To enhance accuracy for hand detection in arbitrary position of the image, we propose attention-based re-detection. After roughly detection, we translate the image to centralize detected hand before we re-detect hand precisely. Moreover, the size of image input into CNN remains the same since we fill those part exceeding the original image with mean values to avoid feature response.

This pipeline is, from bionic perspective, similar to human's visual action. According to vision researches, human being unconsciously focuses on the central part of vision, which academically named "center bias" [14, 15] or "center prior" [20, 21]. While gazing at target object, human eyes will relocate target by moving it to vision center. This incurs our Attention-based Hand Detector.

We also fine-tune the CNN parameters with synthesized training samples roughly centralized on hand with a random location bias. This significantly improves the detection accuracy by 10.9% of overlap rate.

*3.1.3 Mechanism behind AHD*

There are mainly two reasons for AHD's impressive performance.

Firstly, AHD reduces the impact of unbalanced training data as Figure 3 shows. Trained on this unbalanced dataset, the CNN is good at detecting hands in Attention Focus Zone but performs poorly while detecting hands in Surrounding Zone. However, in the pipeline of AHD, hand will be roughly translated to the center of the image and therefore achieves better accuracy by re-detection.

Secondly, the fine-tuning of AHD actually conducts re-learning in lower dimension feature space, which demands less training data and minor capacity of neural networks. To locate hand, the fully connected layer, as a regressor, has to build a mapping from a high dimension ( height * width * channel ) feature space of last convolutional layer to a 4-dimension space of hand bounding box coordinates. Fortunately, we found that target hand region has certain responses in the corresponding area of feature maps, owing to CNN's shared parameters [27]. As a result, the feature patches about hand occur in different locations as the hand moves around. Once we translate the image to centralize roughly detected hand, the feature responses about hand will roughly locate at the center of the feature maps. This lowers dimension of feature space, which reduces the complexity of the mapping by regressor. Besides, after fine-tuning the CNN adapts to the sharp edge caused by filling mean values.

More importantly, this attention-based detection pipeline can be easily extended to other class-specific object detection and even general object detection because it effectively deals with the unbalance of training data and therefore reduces the training data needed. Moreover this pipeline demands less capacity of fully connected layer.

### 3.2. Multi-point Fingertip Detector: MFD

Taking bounding box detected by AHD as input, the Multi-point Fingertip Detector (MFD) locates the key points in the bounding box without the complex disturbance in the image background. We design the network as Figure 5.

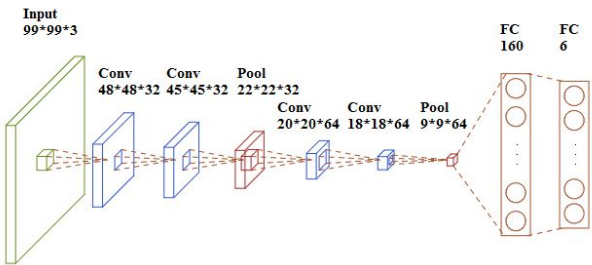

Figure 5. Multi-point Fingertip Detector network

According to related works [22, 23], researchers usually describe a hand with multi-points model, indicating the fact that multiple key points of hand form spatial constraints. For instance, index fingertip location is biologically adjacent to index finger joint. Shuffling, swaying, twisting or flipping of hand will indeed never change the structural constraint among hand points. With such prior knowledge, CNN could learn structural features of hand model. Therefore, we apply multi-point strategy, detecting two key points (index fingertip and index finger joint) of hand simultaneously. Experimental result shows that combined with AHD, such strategy promotes performance by 23% of fingertip error.

### 4. EXPERIMENTS AND DISCUSSIONS

In this section, we will provide the experimental detail as well as result for better illustration of our method.

### 4.1. Data Augmentation

Data augmentation is significant, since it is capable to reduce overfitting. For AHD, we firstly resize the original image of size 640*480 to 256*256 images. Then resized images are randomly cropped to size of 227*227. For MFD, we make random affine transformation of scale and rotation for hand in bounding box. Meanwhile we modify the corresponding label accordingly. Data augmentation enlarges coverage of data, leading to robustness of our algorithm.

### 4.2. Performance of Hand Detection

Besides applying the same network to detect hand in the center of image, we also fine-tune the network to enhance learning on samples whose hand locates in the Attention Focus Zone. In detail, from original training samples, we synthesize many training samples which roughly center on the hand with a random bias between 0 to 50 pixels, and fill the exceeding part with mean values, which forces the corresponding features to be zero.

Figure 6 shows results in different settings: a) rough detection only; b) centralized re-detection; c) centralized re-detection with fine-tuning.

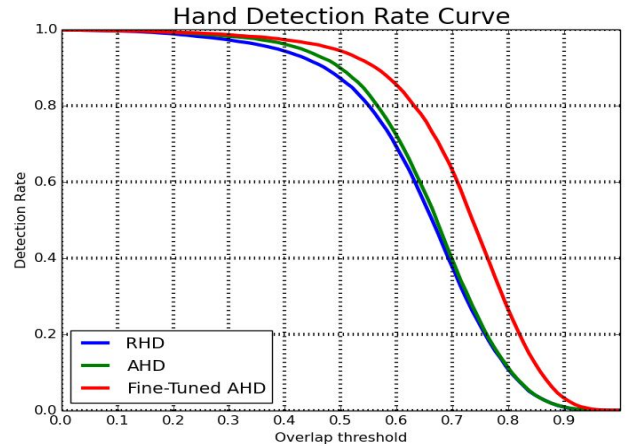

Figure 6. Curve of overlap rate threshold and detection rate, i.e. considering 0.5 overlap rate as threshold, with fine-tuned AHD method 94% validation samples are proved detected.

Respectively, average overlap rates with ground truth of the three settings are 0.642, 0.653, and 0.712. Relatively, the attention pipeline improves the hand detection accuracy for

1.7% in average. Moreover, fine-tuning enormously improves hand detection accuracy for 10.9% in average owing to that fine-tuning enhances the neural network's ability to capture features in the center of the feature maps. It is worth to mention that AHD mainly improves accuracy for those samples whose hand locates in Surrounding Zone while hands in validation dataset mostly locate in Attention Focus Zone as Figure 3 shows.

### 4.3. Performance of Fingertip Detection

Following the output of hand detection module, we perform a cross comparison experiment for attention-based re-detection strategy and multi-point constraint strategy.

*4.3.1 Cross Comparison*

While evaluating the fingertip prediction error, it should be noticed that inner mismatching of cascaded structure would cause larger error. Once first-level hand detector produces an incorrect output bounding box, second-level fingertip detector will accordingly predict poorly, which indicates the importance of accurate hand detection.

Therefore, for hand detection module, we compare three strategies. They are: a) rough hand detector (RHD); b) attention-based hand detector (AHD); c) ground truth hand bounding box (GT). And for the fingertip detection module, Single Point Detector (SPD) is compared with Multi-point Fingertip Detector (MFD).

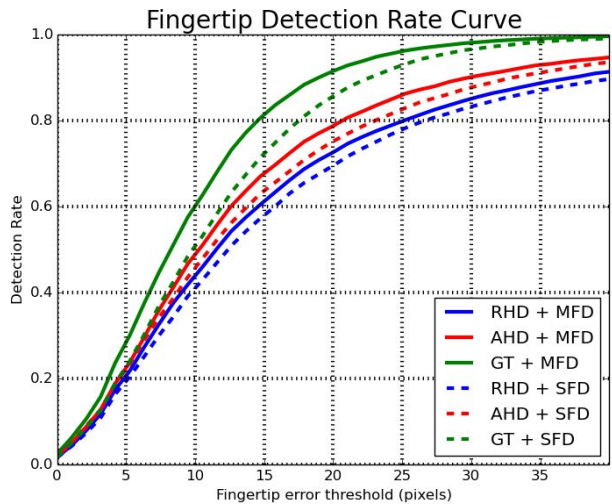

Figure 7. Curve of error threshold and detection rate, i.e. considering 20 pixels as threshold, with AHD+MFD method, 79% of validation samples are proved successfully detected.

Respectively, average fingertip detection error for RHD+MFD, AHD+MFD, GT+MFD, RHD+SFD, AHD+SFD, GT+SFD methods are 18.93, 15.71, 10.71, 20.34, 16.93 and 12.50 pixels. It confirms the effectiveness of AHD and MFD. The proposed AHD-MFD method enormously improves the fingertip detection accuracy for 23%.

*4.3.2 Real-time performance*

Detection methods based on sliding windows and object proposals encounter computing bottleneck. Researchers begin to reformulate detection as a direct regression problem. However, different from YOLO [24], our attention-based work indicates that "Look Twice" is a better tradeoff between accuracy and speed. Applying CAFFE framework [25] on single GPU GTX 980, our system achieves real-time detection. The time consumption of AHD is about 5.76 ms, while MFD consumes about 0.68 ms. Combining extra time on image processing, the overall pipeline costs about 9.65 ms in total.

*4.3.3 Result of challenging instances*

Finally some impressive instances are shown in the Figure 8. These instances show effectiveness and robustness of our proposal even if the image frame is challenging.

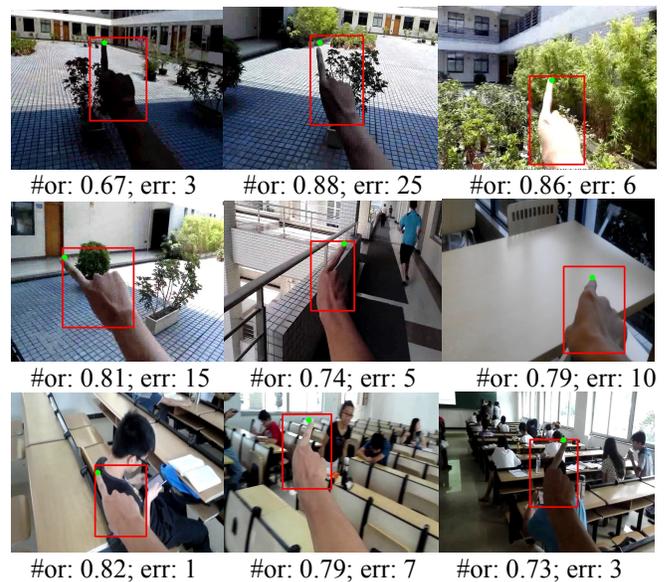

#or: 0.67; err: 3   #or: 0.88; err: 25   #or: 0.86; err: 6
#or: 0.81; err: 15  #or: 0.74; err: 5    #or: 0.79; err: 10
#or: 0.82; err: 1   #or: 0.79; err: 7    #or: 0.73; err: 3

Figure 8.Samples of validation (or: overlap rate, err: fingertip error), covering dark, light, indoor, outdoor environments, hand shape varying and motion blur.

### 5. CONCLUSION

This paper presents a cascaded CNN pipeline including two level modules. The first-level CNN is cultivated by visual attention prior, named Attention-based Hand Detector, and the second-level CNN is based on spatial constraint of hand, named Multi-point Fingertip Detector. Training and validating on the dataset we built, which covers various challenges, the AHD-MFD integrated pipeline performs well, that the hand detector achieves 0.71 average overlap rate on hand bounding box detection, and integrated pipeline achieves 15.71 average error out of 640*480 video sequences. Future work will focus on multiple hands detection and tracking.